\pdfoutput=1
\documentclass[11pt]{article}

\usepackage[final]{acl}

\usepackage{times}
\usepackage{latexsym}

\usepackage[T1]{fontenc}
\usepackage[utf8]{inputenc}

\usepackage{microtype}

\usepackage{inconsolata}

\usepackage{amsmath}
\usepackage{array}

\usepackage{graphicx}
\usepackage{algorithm}
\usepackage{algorithmic}
\usepackage{booktabs}
\usepackage{multirow}
\usepackage{amsfonts}
\usepackage{amssymb}

\usepackage{nicefrac}
\usepackage{url}
\usepackage{xcolor}
\usepackage{colortbl}
\usepackage{makecell}

\makeatletter
\newcommand{\ssymbol}[1]{$^{\@fnsymbol{#1}}$}
\makeatother

\newcommand{\mysubparagraph}[1]{\textit{#1}\hspace{1.8ex}}

\newcommand{\mytabfootnotesize}{\fontsize{5}{6}\selectfont}
\newcommand{\myfootnotesize}{\fontsize{6}{7}\selectfont}
\newcommand{\mypromptfootnotesize}{\fontsize{5}{6}\selectfont}
\newcommand{\myauthorsize}{\fontsize{11}{12}\selectfont}

\title{Large Language Models in the Clinic: A Comprehensive Benchmark}

\author{\myauthorsize Fenglin Liu\textsuperscript{1}, Zheng Li\textsuperscript{2}\thanks{\ \ Corresponding authors.
\\
\indent \ \ \ \tt \{liufengl, amzzhe\}@amazon.com}, Hongjian Zhou\textsuperscript{1}, Qingyu Yin\textsuperscript{2}, Jingfeng Yang\textsuperscript{2}, Xianfeng Tang\textsuperscript{2},  \\ \myauthorsize \bf  Chen Luo\textsuperscript{2}, Ming Zeng\textsuperscript{2}, Haoming Jiang\textsuperscript{2}, Yifan Gao\textsuperscript{2},  Priyanka Nigam\textsuperscript{2}, Sreyashi Nag\textsuperscript{2}, Bing Yin\textsuperscript{2},   \\ \myauthorsize \bf Yining Hua\textsuperscript{3}, Xuan Zhou\textsuperscript{4},  Omid Rohanian\textsuperscript{1}, Anshul Thakur\textsuperscript{1}, Lei Clifton\textsuperscript{5}, David A. Clifton\textsuperscript{1,6}\footnotemark[1] 
\\
\myauthorsize \textsuperscript{1} Institute of Biomedical Engineering, Department of Engineering Science, University of Oxford     \\
\myauthorsize \textsuperscript{2} Amazon
\textsuperscript{3} Harvard T.H. Chan School of Public Health
\textsuperscript{4} Institut polytechnique de Paris \\
\myauthorsize \textsuperscript{5} Nuffield Department of Population Health, University of Oxford\\
\myauthorsize \textsuperscript{6} Oxford-Suzhou Centre for Advanced Research, Suzhou, China \\
}

\begin{document}
\maketitle

\begin{abstract}

The adoption of large language models (LLMs) to assist clinicians has attracted remarkable attention. Existing works mainly adopt the close-ended question-answering (QA) task with answer options for evaluation. However, many clinical decisions involve answering open-ended questions without pre-set options. To better understand LLMs in the clinic, we construct a benchmark \textit{ClinicBench}. We first collect eleven existing datasets covering diverse clinical language generation, understanding, and reasoning tasks. Furthermore, we construct six novel datasets and clinical tasks that are complex but common in real-world practice, e.g., open-ended decision-making, long document processing, and emerging drug analysis. We conduct an extensive evaluation of twenty-two LLMs under both zero-shot and few-shot settings. Finally,  we invite medical experts to evaluate the clinical usefulness of LLMs
\footnote{The benchmark data is available at \url{https://github.com/AI-in-Health/ClinicBench}.}.

\end{abstract}

\section{Introduction}

Large language models (LLMs), such as ChatGPT \cite{OpenAI2023GPT4TR}, are increasingly being recognized for their potential in healthcare to aid clinical decision-making.
Recently, many efforts have been made to develop medical LLMs \cite{zhou2023survey,singhal2023large,singhal2023towards,liu2023medical}.
Existing research shows that medical LLMs outperform human experts across a variety of medical tasks.
In particular, MedPrompt \cite{nori2023can} and MedPaLM-2 \cite{medpalm2} have respectively achieved a competitive accuracy of 90.2 and 86.5 compared to human experts 87.0 \cite{wu2023pmc} on the United States Medical Licensing Examination (USMLE).
\nocite{liu2021contrastive}

\begin{figure}[t]
\centering
\includegraphics[width=1\linewidth]{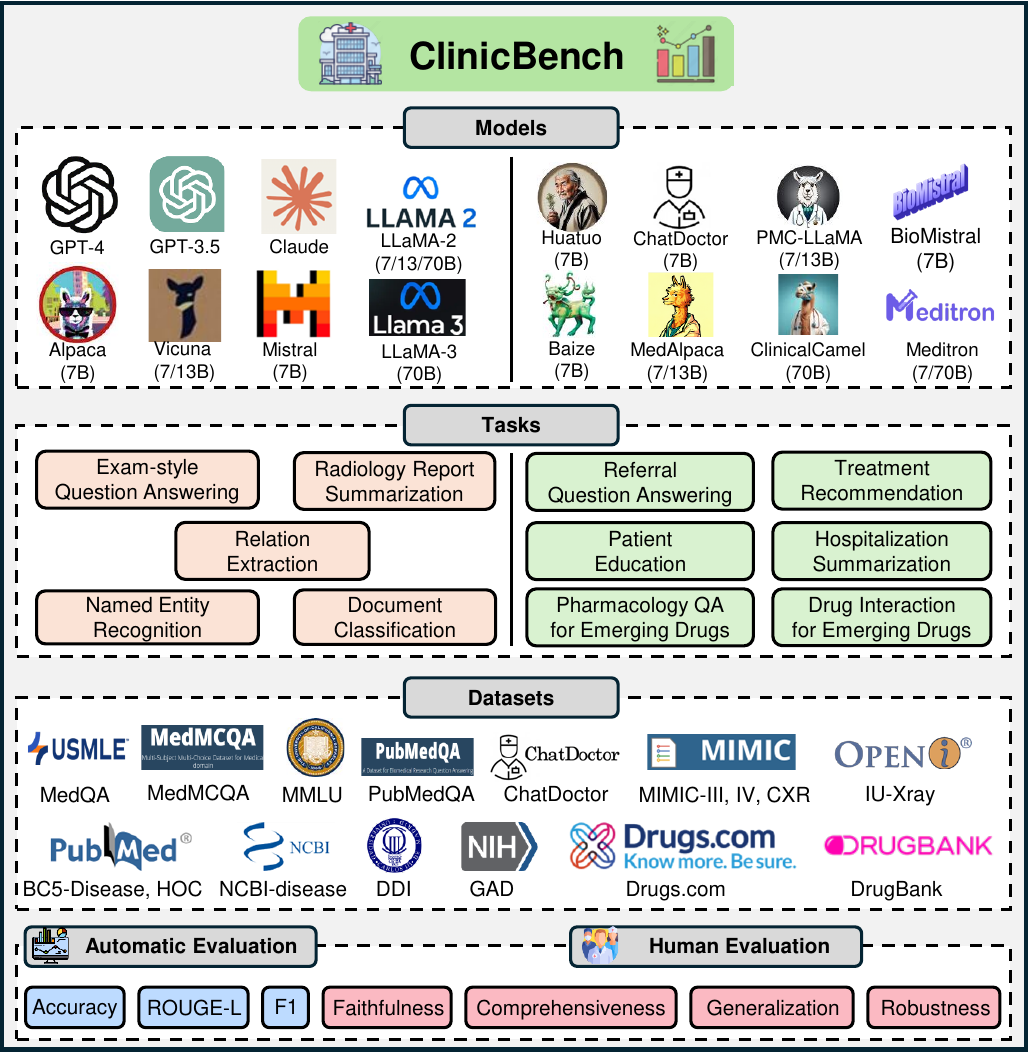}
\caption{Overview of our ClinicBench, which includes 22 LLMs, 11 tasks, 17 datasets, and multiple metrics across automatic and human evaluations.}
\label{fig:benchmark}
\end{figure}

Despite the promising results of existing LLMs, several issues need to be addressed for better use of LLMs in assisting clinicians:
\textbf{(i) Limited evaluation:} Most works only focus on evaluating LLMs in close-ended (exam-style) QA tasks, overlooking their evaluation in other scenarios, such as clinical language understanding and generation \cite{he2023survey};
\textbf{(ii) Limited task:} Current works primarily focus on non-clinical machine learning tasks, which cannot adequately evaluate the models' ability to solve complex clinical problems, e.g., health education \cite{safranek2023role}, treatment recommendation \cite{wilhelm2023large}, and emerging drug analysis;
\textbf{(iii) Limited comparison:}  Most works either provide limited qualitative examples \cite{patel2023chatgpt} or use limited baselines (mainly focusing on ChatGPT) for quantitative comparisons \cite{chen2023extensive,chen2023large,jahan2024comprehensive}.
Therefore, existing works fall short of providing a thorough comparative analysis of different LLMs across diverse clinical scenarios and tasks.

In this paper, as shown in Figure~\ref{fig:benchmark}, we propose the ClinicBench, which encompasses eleven downstream tasks (across three different scenarios, i.e., reasoning, generation, and understanding) and seventeen datasets to provide a comprehensive evaluation of LLMs in the clinic.
Most recently, several works \cite{chen2023extensive,chen2023large} have attempted to benchmark LLMs in healthcare. However, they only adopt the non-clinical machine learning tasks from existing benchmarks (i.e., BLUE \cite{Peng2019BLEUBERT} and BLURB \cite{gu2021domain}) for evaluation.
As shown in Table~\ref{tab:ClinicBench}, we further propose six clinical tasks and build six novel datasets to evaluate the performance of LLMs in multiple complex clinical scenarios, i.e., open-ended decision-making, long document processing, and new drug understanding, all of which are very common in the clinic.
Meanwhile, it is worth noting that existing works mainly use three LLMs, i.e., LLaMA, GPT-3.5, and GPT-4, for evaluation. In our work, as shown in Table~\ref{tab:baselines}, we collect twenty-two diverse LLMs for a comprehensive comparison.
In terms of evaluations, existing works mainly consider 0-shot and 1-shot experiments. We further perform 3-shot and 5-shot experiments, and provide human evaluations to offer insights into LLMs in the clinic.

Finally, to build medical LLMs, we notice that existing works \cite{he2023survey} mainly adopt medical consultant dialogues, exam-style QA, and articles as fine-tuning data, which, however, cannot directly or fully represent clinical knowledge. For example, dialogues contain irrelevant and incomplete information, and articles may focus on laboratory results rather than clinical practice.
Thus, we explore using clinical-standard knowledge bases collected from the clinic to develop a medical LLM. We then analyze how different types of fine-tuning data affect medical LLMs' performance. 
The evaluation proves the importance of introducing clinical-standard knowledge bases for fine-tuning and increasing the diversity of fine-tuning data.

The main contributions of this work are:
\begin{itemize}
    \item We construct the \textit{ClinicBench} with 3 scenarios, 11 tasks, and 17 datasets, including over 20,000 test samples, to benchmark 22 LLMs under both zero-shot and few-shot settings.

    \item We build 6 novel datasets tailored to clinical practice to measure the capabilities of LLMs in solving complex but very common clinical problems: open-ended decision-making, long document processing, and new drug analysis. 

    \item We further perform a human evaluation to benchmark the clinical usefulness of LLMs. We conduct a preliminary exploration of using clinical-standard knowledge bases as the fine-tuning data to develop medical LLMs and analyze the effect of the fine-tuning data.

\end{itemize}

In the following sections, we introduce the details of our benchmark in Section~\ref{sec:clinicbench}. We report the main results of our benchmark in Section~\ref{sec:automatic}. We discuss the differences in LLMs' performance on machine learning and clinical tasks in Section~\ref{sec:clinical_task}. We analyze the effect of few-shot learning of LLMs in medical tasks in Section \ref{sec:few-shot}. We evaluate the clinical usefulness of LLMs in Section \ref{sec:human-evaluation}. Finally, we analyze the effect of medical IFT data in Sections \ref{sec:IFT} and \ref{sec:example_IFT}.

\begin{table*}[t]
\centering
\myfootnotesize
\setlength{\tabcolsep}{3pt}
 
\begin{tabular}{llllll}
\toprule
Scenarios & Tasks & Descriptions & Datasets   & Sizes & Metrics \\
\midrule 
\multirow{9}{*}{\begin{tabular}[c]{@{}l@{}} Clinical \\ Language \\ Reasoning \end{tabular}} & \multirow{4}{*}{\begin{tabular}[c]{@{}l@{}} Exam-style   \\ QA \end{tabular}} &  \multirow{4}{*}{\begin{tabular}[c]{@{}l@{}} Predict the correct answer to the given question from the provided \\  choices.  \end{tabular}} & MedQA (USMLE) \cite{medqa}   & 1,273 & Accuracy
\\
& && MedMCQA \cite{Medmcqa}   & 4,183 & Accuracy 
\\ 
& && MMLU-Med \cite{hendrycks2020measuring}  & 272 & Accuracy 
\\
& && PubMedQA  \cite{PubmedQA}   & 500 & Accuracy 
\\ \cmidrule(l){2-6}

& \cellcolor{gray!10}\begin{tabular}[c]{@{}l@{}} Referral  \\ QA  \end{tabular} & \cellcolor{gray!10}\begin{tabular}[c]{@{}l@{}} Predict the correct answer to the given question about patients' \\ treatments and medications, based on their referral letters. \end{tabular}  & \cellcolor{gray!10}Derived from MIMIC-IV \cite{johnson2023mimiciv} & \cellcolor{gray!10}1,057 & \cellcolor{gray!10}Accuracy \\ \cmidrule(l){2-6}

& \cellcolor{gray!10}\begin{tabular}[c]{@{}l@{}} Treatment \\ Recommendation  \end{tabular} & \cellcolor{gray!10}\begin{tabular}[c]{@{}l@{}} Recommend all appropriate drugs for the treatment of patients,  \\  given their conditions and symptoms. \end{tabular} & \cellcolor{gray!10}Derived from ChatDoctor \cite{li2023chatdoctor} & \cellcolor{gray!10}796 & \cellcolor{gray!10}F1
\\ \midrule 

\multirow{9}{*}{\begin{tabular}[c]{@{}l@{}} Clinical \\ Language \\ Generation \end{tabular}} & \multirow{2}{*}{\begin{tabular}[c]{@{}l@{}} Radiology Report \\ Summarization \end{tabular}} &  \multirow{2}{*}{\begin{tabular}[c]{@{}l@{}} Generate a concise 'Impression' section from the lengthy  'Findings' \\ section in a radiology report. \end{tabular}} & MIMIC-CXR \cite{johnson2019mimic} & 3,269 & ROUGE-L
 \\
& & & IU-Xray  \cite{Dina2016IU-Xray}  & 341 & ROUGE-L

\\  \cmidrule(l){2-6}
&  \cellcolor{gray!10}\begin{tabular}[c]{@{}l@{}} Hospitalization \\ Summarization \end{tabular}  & \cellcolor{gray!10}\begin{tabular}[c]{@{}l@{}} Summarize the key diagnostic information and significant results \\ based on the patients' multiple health (\textit{long}) records during \\ hospitalization, e.g., physician notes, nursing notes, and medication.\end{tabular}  &\cellcolor{gray!10}Derived from MIMIC-IV \cite{johnson2023mimiciv}  & \cellcolor{gray!10}382 &\cellcolor{gray!10}ROUGE-L

\\  \cmidrule(l){2-6}
&  \cellcolor{gray!10}\begin{tabular}[c]{@{}l@{}} Patient Education  \end{tabular}  & \cellcolor{gray!10}\begin{tabular}[c]{@{}l@{}} Generate educational instructions to help patients manage their \\ conditions, according to their health (\textit{long}) documents.\end{tabular}  &\cellcolor{gray!10}Derived from MIMIC-III \cite{Johnson2016MIMICIII}  & \cellcolor{gray!10}181  &\cellcolor{gray!10}ROUGE-L

\\ \midrule 

\multirow{13}{*}{\begin{tabular}[c]{@{}l@{}} Clinical \\ Language \\ Understanding \end{tabular}} & \multirow{2}{*}{\begin{tabular}[c]{@{}l@{}} Named Entity \\ Recognition (NER) \end{tabular}}  &  \multirow{2}{*}{\begin{tabular}[c]{@{}l@{}} Extract medical entities mentioned in clinical notes and classify them \\ according to relevant symptoms, medications, dosages, and procedures. \end{tabular}} & BC5-Disease \cite{li2016biocreative}   & 4,797 & F1 entity-level 
\\
&&  &  NCBI-Disease \cite{dougan2014ncbi}   & 940 & F1 entity-level \\ \cmidrule(l){2-6}
& \multirow{2}{*}{\begin{tabular}[c]{@{}l@{}} Relation \\ Extraction (RE) \end{tabular}} &  \multirow{2}{*}{\begin{tabular}[c]{@{}l@{}} Identify the relations, e.g., the mechanism of interaction, the effect of \\ interaction, between medical entities mentioned in the text. \end{tabular}} &  DDI \cite{segura2013semeval}   & 5,716 & Micro F1 \\
& & & GAD \cite{becker2004genetic}   & 534 & Micro F1 \\ \cmidrule(l){2-6}
&  \begin{tabular}[c]{@{}l@{}} Document \\ Classification (DC) \end{tabular} &   {\begin{tabular}[c]{@{}l@{}} Predict multiple correct labels to the input clinical document. \end{tabular}} &  HoC \cite{baker2016automatic}   & 315 & Micro F1 \\ \cmidrule(l){2-6}

&  \cellcolor{gray!10}\begin{tabular}[c]{@{}l@{}} Pharmacology QA \\ for Emerging Drugs \end{tabular}  & \cellcolor{gray!10}\begin{tabular}[c]{@{}l@{}} Predict the correct answer to the given pharmacology question for the \\  new drugs released between October 2023 and April 2024.

\end{tabular}  &\cellcolor{gray!10}Derived from DrugBank \cite{wishart2018drugbank}  & \cellcolor{gray!10}213   &\cellcolor{gray!10}Accuracy 
\\ \cmidrule(l){2-6}

&  \cellcolor{gray!10}\begin{tabular}[c]{@{}l@{}} Drug Interaction \\ for Emerging Drugs \end{tabular}  & \cellcolor{gray!10}\begin{tabular}[c]{@{}l@{}} Assess whether the therapeutic efficacy of the Moderna COVID-19 \\ Vaccine can be decreased when used in combination with other drugs.
\end{tabular}  &\cellcolor{gray!10}Derived from Drug.com \cite{drugs} & \cellcolor{gray!10}200  &\cellcolor{gray!10}Accuracy
\\
\bottomrule
\end{tabular}
\caption{Overview of our evaluation scenarios, which includes eleven existing datasets covering five non-clinical machine learning tasks and six novel datasets covering six complex clinical tasks (gray-highlighted text).} 
\label{tab:ClinicBench}
\end{table*}

\section{Key Findings}

For clarity, we summarize the main findings from our benchmark as follows:

\noindent $\bullet$  \textbf{Commercial LLMs}: Closed-source commercial LLMs, especially GPT-4, outperform all existing open-source public LLMs on all tasks and datasets.

\noindent $\bullet$ \textbf{State-of-the-art (SOTA)}: 
LLMs achieve superior performance only on exam-style QA tasks with provided options, competing with human experts and substantially outperforming previous task-specific SOTA methods. However, LLMs perform poorly in open-ended decision-making, generation, and understanding.

\noindent $\bullet$ \textbf{Medical LLMs}: Fine-tuning LLMs on medical data can improve their reasoning and understanding of medical data, but it may decrease their summarization ability. In-domain fine-tuning \cite{van2024adapted} and few-shot prompting could be potential solutions to address this limitation.

\noindent $\bullet$ \textbf{Clinical Tasks}: Existing LLMs are less effective in dealing with complex clinical tasks, demonstrating a significant drop in performance. Nevertheless, commercial LLMs drop slightly less compared to public LLMs. Medical LLMs can adapt better to clinical tasks compared to general LLMs.

\noindent $\bullet$ \textbf{Few-shot Learning}: It leads to better reasoning and generation performance (i.e., 1-shot or 3-shot learning achieves the best reasoning performance and more shots consistently lead to better generation performance), but impairs the understanding performance of LLMs.

\noindent $\bullet$  \textbf{Clinical usefulness}: Medical LLMs produces more factual and safe responses than general LLMs, but perform worse in generating complete and user-preferred responses. A certain degree of hallucination may offer benefits to clinicians by providing a broader spectrum of diagnostic suggestions, which could be advantageous in the diagnosis of rare diseases.

\noindent $\bullet$  \textbf{Instruction Fine-tuning}: Different types of IFT data bring improvements from different aspects; more diverse IFT data can lead to better medical LLMs, highlighting the importance of improving the diversity of IFT data, which is as crucial as increasing the quantity of training data.

Overall, our results show that the close-ended QA task is the major task in which current LLMs can outperform state-of-the-art task-specific models and are comparable to human experts. However, clinical decisions often confront open-ended questions that lack pre-determined answer choices.
Our results further reveal that current LLMs' performance drops clearly when applied to open-ended decision-making, long document processing, and new drug understanding.
We hope that this work can offer a holistic view of LLMs in healthcare, aiming to bridge the current gaps and advance the integration of LLMs in clinical applications.

\section{ClinicBench}
\label{sec:clinicbench}
Table~\ref{tab:ClinicBench} illustrates our benchmark. Here, we mainly introduce our built six clinical tasks and datasets.
Please refer to our supplementary material and Appendix~\ref{appendix:ml-tasks} for more details of our benchmark.

i) \textit{Referral QA}: When a patient returns to their GP or is referred to another hospital, this task can help clinicians quickly understand the patient's treatment and medication.
We randomly extract 1,000 referral letters from MIMIC-IV \cite{johnson2023mimiciv} and use GPT-4 to generate multiple QA pairs about medications and treatments for each letter. After review, correction, and filtering by experts from our university's health department, we finally obtain 1,057 Q\&A pairs.

ii) \textit{Treatment Recommendation}: It requires providing all possible appropriate medications for treating the current patient's condition, thus helping the clinician designate a treatment plan.
We collect 796 patient-physician conversations, which describe the patient's diseases and symptoms, from ChatDoctor \cite{li2023chatdoctor}. We further collect the corresponding drug recommendations given by the physicians.
During our evaluation, we ask the LLMs to list all available drugs for the treatment of patients based on their diseases or symptoms.

iii) \textit{Hospitalization Summarization}: Clinicians need to spend about 40\%-50\% of their time in their daily work \cite{sinsky2016allocation} reading lots of patients' health documents and writing a hospitalization summary that highlights key diagnostic information, which is crucial for the patient's discharge or transfer to another hospital.
To this end, we construct 382 pairs of clinical documents and summaries from MIMIC-IV resource \cite{johnson2023mimiciv}. The average length of the input documents is around 1,675 words.

iv) \textit{Patient Education}: Similarly, clinicians need to read lots of health documents to write educational materials to guide patients on how to better manage their conditions. Therefore, a desirable patient education generation system can substantially reduce clinical workload.
We adopt the MIMIC-III \cite{Johnson2016MIMICIII} to collect 181 pairs of health documents and educational instructions.
The average length of the input documents is 3,037 words.

v) \textit{Pharmacology QA for Emerging Drugs:} 
It aims to answer pharmacology-related questions based on the given new drugs.
We collect 213 new drugs released on the pharmaceutical knowledge database DrugBank \cite{wishart2018drugbank} between October 2023 and April 2024. Then, we use GPT-4 \cite{gpt-4} to generate 213 question-answer pairs, which are further reviewed by experts.

vi) \textit{Drug Interaction for Emerging Drugs}: It aims to predict the effects of known drug combinations given the descriptions of new drugs. 
To this end, we randomly chose 100 identified drug interactions with the Moderna COVID-19 Vaccine (2023-2024 Formula) and 100 drugs without interaction from  Drug.com \cite{drugs}.
The LLMs are tasked with assessing whether the therapeutic efficacy of the vaccine can be decreased when used in combination with other drugs. 

\paragraph{Discussion}
Unlike exam-style QA tasks with answer options, the treatment recommendation is an open-ended task where the LLMs need to rely on their own knowledge to reason and make decisions;
Both patient education and hospitalization summarization tasks require LLMs to process patient documents with lengths around 2,000-3,000 words, and thus can evaluate LLMs' ability to deal with \textit{long} health documents, which are common in the clinic;
The tasks of pharmacology QA and drug interaction for emerging drugs are crucial for supporting the decision-making and management of new drugs, which frequently emerge in real-world clinical practice.

\begin{table}[t]
\centering
\scriptsize
\setlength{\tabcolsep}{3pt}
 
\begin{tabular}{clcc}
\toprule
\textbf{Types} & \textbf{Methods} & \multicolumn{1}{c}{\textbf{\# Params}}   \\
\midrule 
\multirow{13}{*}{\rotatebox{90}{\begin{tabular}[c]{@{}c@{}} General \\ Large Language Models \end{tabular}}} 
&  Claude-2 \cite{Claude2} & {Commercial}   \\
&  GPT-3.5-turbo \cite{openai2023chatgpt} &  Commercial  \\
&  GPT-4-0613 \cite{gpt-4} & Commercial \\
\cmidrule{2-3}
& Alpaca \cite{alpaca} &  7B    \\ 
& Vicuna-7B \cite{vicuna} &  7B    \\ 
& LLaMA-2-7B \cite{touvron2023llama2} &  7B    \\ 
& Mistral \cite{jiang2023mistral} &  7B  \\

\cmidrule{2-3}
& Vicuna-13B \cite{vicuna} &  13B    \\ 
& LLaMA-2-13B \cite{touvron2023llama2}& 13B \\  
\cmidrule{2-3}

& LLaMA-2-70B \cite{touvron2023llama2}  & 70B \\
& LLaMA-3-70B \cite{llama3}  & 70B \\
\midrule
\multirow{13}{*}{\rotatebox{90}{\begin{tabular}[c]{@{}c@{}} Medical \\ Large Language Models \end{tabular}}} 
& Huatuo \cite{zhang2023huatuogpt} &  7B    \\ 
& ChatDoctor \cite{li2023chatdoctor}&  7B    \\ 
& PMC-LLaMA-7B \cite{wu2023pmcllama}&  7B    \\ 
& Baize-Healthcare \cite{xu2023baize} &  7B    \\ 
& MedAlpaca-7B \cite{han2023medalpaca} &  7B    \\ 
& Meditron-7B  \cite{chen2023meditron} &  7B    \\ 
& BioMistral \cite{labrak2024biomistral} &  7B    \\ 
\cmidrule{2-3}
& PMC-LLaMA-13B \cite{wu2023pmcllama}&  13B    \\ 
& MedAlpaca-13B \cite{han2023medalpaca} & 13B \\
\cmidrule{2-3}
& ClinicalCamel \cite{toma2023clinical}& 70B \\
& Meditron-70B  \cite{chen2023meditron} & 70B \\

\bottomrule
\end{tabular}
\caption{We collect 22 LLMs (i.e., 11 general LLMs and 11 medical LLMs) covering open-source public LLMs and closed-source commercial LLMs, across different numbers of parameters from 7 to 70 billion (B).}
\label{tab:baselines}
\end{table}

\begin{table*}[t]
\centering
\mytabfootnotesize
\setlength{\tabcolsep}{2.5pt}
 
\begin{tabular}{clcccccccccccccccccc}
\toprule
\multirow{3}{*}[-6pt]{\textbf{Types}} & \multirow{3}{*}[-6pt]{\textbf{Methods}} & \multirow{3}{*}[-6pt]{\textbf{\# Params}} & \multicolumn{6}{c} {\textbf{Clinical Language Reasoning}} & \multicolumn{4}{c} {\textbf{Clinical Language Generation}} & \multicolumn{7}{c} {\textbf{Clinical Language Understanding}} 
\\ \cmidrule(lr){4-9} \cmidrule(lr){10-13}  
\cmidrule(lr){14-20}

&&& \multicolumn{4}{c} {Exam-style QA} & \multirow{2}{*}[-1.5pt]{\begin{tabular}[c]{@{}c@{}} \cellcolor{gray!10}Referral \\ \cellcolor{gray!10}QA \end{tabular}} & \multirow{2}{*}[-1.5pt]{\begin{tabular}[c]{@{}c@{}} \cellcolor{gray!10} Treat  \\ \cellcolor{gray!10} Recom. \end{tabular}}

&  \multicolumn{2}{c} {Report Summari.} & \multirow{2}{*}[-1.5pt]{\begin{tabular}[c]{@{}c@{}} \cellcolor{gray!10} Hospitaliz.  \\ \cellcolor{gray!10} Summari.  \end{tabular}} & \multirow{2}{*}[-1.5pt]{\begin{tabular}[c]{@{}c@{}} \cellcolor{gray!10} Patient   \\ \cellcolor{gray!10} Education \end{tabular}}  &  \multicolumn{2}{c} {NER} &  \multicolumn{2}{c} {RE} &  DC  & \multirow{2}{*}[-1.5pt]{\begin{tabular}[c]{@{}c@{}} \cellcolor{gray!10} Pharma. \\ \cellcolor{gray!10} QA  \end{tabular}}   & \multirow{2}{*}[-1.5pt]{\begin{tabular}[c]{@{}c@{}} \cellcolor{gray!10} Drug \\ \cellcolor{gray!10} Inter.   \end{tabular}}
\\ \cmidrule(lr){4-7} \cmidrule(lr){10-11} \cmidrule(lr){14-15} \cmidrule(lr){14-15} \cmidrule(lr){16-17} \cmidrule(lr){18-18} 

&&& MedQA &   MedMCQA  &   MMLU  &  PubMedQA  & & & MIMIC & IU-Xray  & & & BC5  & NCBI &  DDI&  GAD & HoC \\ \midrule
\multicolumn{2}{l}{Task-specific SOTA} &  - &  44.6 & 43.0 & - & 60.2 & - & - & \textbf{46.1} & \textbf{67.9}  & - & - & \textbf{90.0} & \textbf{89.4} &  \textbf{84.1} & \textbf{84.0} & \textbf{85.1} & - & -
\\ \midrule

\multirow{14}{*}{\rotatebox{90}{\begin{tabular}[c]{@{}c@{}} General \\ Large Language Models \end{tabular}}} 
&  Claude-2   &  Com.   &  65.1	 & 60.3	&  78.7 & 	 70.8  & 80.5   & 9.1  & 13.3 & 9.4 & 11.3  &  8.4& 52.9 & 44.2	 &50.4 & 50.7 & 70.8	 & 60.6 & 51.5  \\
&  GPT-3.5-turbo  & Com.  & 61.2	 &  59.4	 &  73.5	 &  70.2 & 81.1   & 7.3  & 14.1 & 10.3 & 10.5 &  9.2  & 52.3 & 46.1 & 49.3 &	 50.8 &	 66.4  & 57.3	& 47.0 \\
&  GPT-4 & Com.   &  \textbf{83.4} &	  \textbf{78.2} 	& \textbf{92.3}	 & \textbf{80.0}	& \textbf{83.2}   & \textbf{18.6}   & 20.7 &  18.6 & \bf  14.2 &  \bf 12.7 & 71.3	 & 58.4	 & 64.6	&68.2  & 83.6  & \bf 63.8   &  \bf 56.5 \\
\cmidrule{2-20}
& Alpaca   & 7B & 34.2	 & 30.1	& 40.8	& 65.2  &  74.8   & 3.5 & 12.6 & 	8.7 & 4.1 & 2.9   &  41.2 & 36.5 & 37.4 & 36.9 & 52.6  & 41.3   & 47.5  \\
& Vicuna-7B  & 7B  & 34.5	& 33.4	& 43.4	& 64.8 &  76.4   & 2.6 & 13.8 & 8.2 & 4.5  &  3.1  &  44.5 & 37.0 & 39.4 & 41.2 & 53.8  &   42.3   & 45.5	\\
& LLaMA-2-7B  & 7B  & 32.9	& 30.6	& 42.3	& 63.4 &  74.5  & 3.3 & 12.3 & 8.6 & 4.9 &  4.6  &  40.1 & 34.8 & 37.9 & 39.3 & 48.6 & 46.5 & 48.0 \\ 
& Mistral  &  7B & 35.7 & 37.8 & 46.3  & 69.4 & 77.7 & 5.0  &   13.2  & 7.9  & 6.1  & 5.3  &  46.8 & 39.9 & 43.5 & 44.3 & 59.6 &  51.2 & 53.0 \\
\cmidrule{2-20}
& Vicuna-13B  &  13B  & 38.0  & 36.4  & 45.6 & 66.2  & 76.8 & 4.6  & 14.5  & 9.4 & 6.2 & 4.7 & 46.2 & 39.0 & 41.3 & 43.5 & 56.7  &  45.1 & 46.0  \\ 
& LLaMA-2-13B  &  13B &  38.1	& 35.5	& 46.0	& 66.8	&  77.1  & 4.8  & 12.0 &  9.1 &  6.4  & 5.6 &  46.6& 38.3 & 39.7 & 41.2 & 55.9  &  46.9  &  47.5 \\ 
\cmidrule{2-20}

& LLaMA-2-70B & 70B & 45.8	& 42.7	& 54.0	& 67.4 &  78.9  & 5.5& 13.9 & 8.0 & 8.3 & 6.8  &  47.8 & 41.5 & 45.6 & 44.7 &  63.2  & 49.3  & 51.5 \\
& LLaMA-3-70B  & 70B & 78.8  & 74.7& 86.4 & 77.4 & 82.4 & 10.2  & 18.4 & 15.5  & 10.9 & 10.1 &  63.7 & 50.2 & 59.7 & 63.1 & 79.0  &  62.4 & 53.0 \\
\midrule
\multirow{14}{*}{\rotatebox{90}{\begin{tabular}[c]{@{}c@{}} Medical \\ Large Language Models \end{tabular}}} 
& Huatuo &  7B &  28.4	  & 24.8 &	 31.6	 & 61.0	&  69.3  & 3.8  & 8.7 & 3.8 & 2.2 &  1.4  & 43.6 & 37.5 & 40.1 & 38.2 & 50.2 &  44.1  & 49.5 \\
& ChatDoctor   &  7B   &  33.2&	 31.5 	& 40.4	& 63.8	&  73.7  &   5.3  & 8.9 & 4.2 & 2.8  & 1.7 &  45.8 & 40.9  & 41.2 & 40.1 & 55.7  &  42.7  & 48.5  \\
& PMC-LLaMA-7B  &  7B  &  28.7 & 29.8  & 39.0   &  60.2  & 70.2 & 4.0 & 7.6  & 4.0 & 3.6 & 1.5 & 45.2 & 37.8 & 40.8 & 42.0 & 55.6 & 45.5  &  51.0 \\ 
& Baize-Healthcare  &  7B&   34.9	& 31.3&	 41.9	& 64.4	&  74.0  & 4.7		& 9.8 &  4.4 & 4.3 &  1.8  &  44.4 & 38.5 & 41.9 & 45.8 & 54.5 & 46.9 &  50.5 \\ 
& MedAlpaca-7B &  7B &    35.1	& 32.9 &	 48.5	& 62.4	&  75.3  & 4.8  & 10.4 & 7.6  & 4.5 & 2.7  & 47.3 & 39.0 & 43.5 & 44.0 & 58.7  & 47.9  & 48.0 \\
& Meditron-7B    &  7B  & 33.5 &   31.1 & 45.2 & 61.6  & 74.9 & 5.8 &  12.5  & 7.8  & 6.8  & 5.9 & 46.5 & 39.2 & 42.7 & 43.3 & 57.9 & 50.7 & 52.0 \\ 
& BioMistral  &  7B  &  35.4 & 34.8 &   52.6 & 66.4 & 77.0 & 7.6 & 14.2 & 8.5 & 7.5 & 6.6 & 48.8 & 40.4 & 46.0 & 48.5 & 64.3 & 54.5 & 54.0 \\ 
\cmidrule{2-20}
& PMC-LLaMA-13B  &  13B  & 39.6 & 37.7 & 56.3 & 67.0 & 77.6  & 4.9 & 9.4  & 5.9  & 4.2 & 2.7 & 51.5 & 43.1 & 48.4 & 48.7 & 65.3  &  48.8  &   51.5  \\ 
& MedAlpaca-13B &  13B   &   37.3	& 35.7	& 51.5 &	 65.6	&  77.4  & 5.1  & 11.7 &  8.6 & 5.0 &  3.5  & 	49.2 & 41.6 & 44.1 & 44.5 & 59.4 & 51.6 & 50.0 \\
\cmidrule{2-20}
& ClinicalCamel & 70B   & 46.4	& 45.8	 &68.4	 &71.0 & 79.8 & 8.4  & 13.0 & 9.6 &  7.9 & 7.2 & 51.2 & 43.7  & 47.6 & 47.2 & 64.8  &  52.6 &  52.5  \\
& Meditron-70B   & 70B &   45.7 & 44.9 & 65.1 & 70.6 &  78.6 & 8.9 & 13.3  & 8.0 &  9.6  & 7.7  & 54.3 & 45.7  &  51.2 & 49.6 &  69.6 & 58.7 & 54.5 \\

\bottomrule
\end{tabular}
\caption{Performance of LLMs under the zero-shot setting.
For comparison, we also report the results of task-specific state-of-the-art (SOTA) models, which are fine-tuned in a fully supervised manner on downstream data and tasks. 
}
\label{tab:benchmark_performance}
\end{table*}
\section{Results}
In this section, we will show the detailed results, analyses, and experimental findings of different LLMs in our benchmark.

\subsection{Settings}
As shown in Table~\ref{tab:baselines}, we collect 22 diverse LLMs to provide a comprehensive benchmark.
Please refer to \citet{zhou2023survey,zhao2023survey,he2023survey} for a detailed introduction to them.
To ensure LLMs achieve optimal performance across different tasks, we use tailored prompts for each task, which are shown in Table~\ref{tab:prompts} of the Appendix.
In detail, for the existing eleven non-clinical tasks, we adopt prompts used in the current state-of-the-art works. 
For each clinical task, we follow previous works \cite{chen2023extensive,jahan2024comprehensive} to design three different prompts and randomly select 100 samples to evaluate their performance using the LLaMA-2-7B, 13B, and 70B models. We then select the best-performing prompt to report the performance of the LLMs on the entire dataset.

\subsection{Automatic Evaluation}
\label{sec:automatic}
We report the results of LLMs in Table~\ref{tab:benchmark_performance} and the results of the task-specific state-of-the-art (SOTA) models from \citet{chen2023extensive,chen2023large,jahan2024comprehensive}.
As we can see, GPT-4 consistently outperforms other LLMs, both general and medical, on all datasets. 
It is worth noting that, on the close-ended exam-style QA task, the three commercial LLMs, i.e., GPT-4, GPT-3.5-turbo, and Claude-2, achieve competitive performance compared to human experts \cite{wu2023pmc}, and substantially outperform previous SOTA. However, this is the only task for which the current LLMs are comparable to the SOTA.
For example, on the clinical language understanding scenario, the best result of LLMs achieved by GPT-4 on the BC5-Disease dataset is 71.3 F1 score, which is far from SOTA (90.0 F1 score).
Overall, all LLMs have a strong reasoning ability to predict accurate answers from the provided options, but perform poorly in other scenarios, particularly in open-ended question decision-making, long document processing, and new drug understanding, as detailed below.

\mysubparagraph{Clinical Language Reasoning}
We can notice that, in terms of open-source public LLMs, medical LLMs consistently achieve better results than general LLMs on all reasoning datasets and across different model sizes, e.g., MedAlpaca-7B, PMC-LLaMA-13B, and ClinicalCamel-70B outperform  LLaMA-2-7B, 13B, 70B models, respectively.
It shows that fine-tuning the general LLMs on medical data improves their performances.
However, on the open-ended task, treatment recommendation, all LLMs achieve poor F1 scores (<20\%), which indicates a considerable need for advancement before LLMs can be integrated into the actual clinical decision-making process without answer options.

\mysubparagraph{Clinical Language Generation}
It is clearly shown that there are significant gaps between the SOTA and LLM performances. In particular, on the tasks of hospitalization summarization and patient education, which involve input documents containing around 2,000-3,000 words, these LLMs are unable to effectively process long clinical documents to achieve desirable performance. This capability is important for understanding a wide variety of medical documents in clinical settings.

\mysubparagraph{Clinical Language Understanding}
Existing LLMs fail to comprehend medical texts, which may be attributed to the lack of necessary extensive expert knowledge, such as medical terminologies and the medical relations between drugs, conditions, and symptoms.
Nevertheless, medical LLMs have better clinical language understanding than general LLMs.
For example, Meditron-70B even outperforms commercial LLMs, Claude-2 and GPT-3.5, in most cases.
When dealing with new drugs, these LLMs exhibit poor performance. On the drug interaction task, where we build a balanced dataset with a 1:1 ratio of positive to negative samples, the performances of nearly half of the LLMs are even \textit{worse than Random} (50\% accuracy).
It indicates that current LLMs are incapable of dealing with new drugs that frequently emerge in the clinic.

\begin{figure}[t]
\centering
\includegraphics[width=1\linewidth]{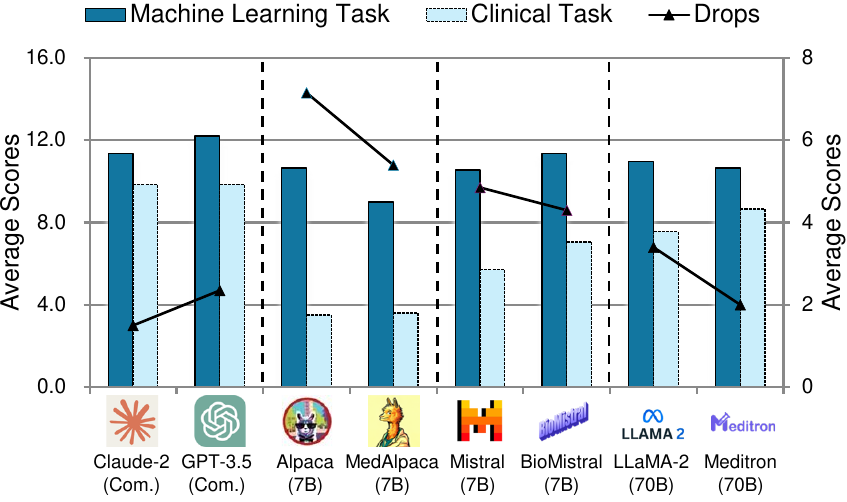}
\caption{Comparison of LLMs' performance on machine learning and clinical tasks. When applied to clinical tasks, the performance drops of the LLMs are shown with the solid line and the right y-axis. Lower is better.}
\label{fig:clinical_task_analysis}
\end{figure}

\subsection{Clinical Task Analysis}
\label{sec:clinical_task}

Table~\ref{tab:benchmark_performance} shows that current LLMs are comparable to SOTA models and human experts on the exam-style QA task.
However, real-world open clinical practice diverges far from the structured nature of exam-taking. 
This paradigm shift from a controlled test environment to the unpredictable and subtle domain of patient care challenges the LLMs, demanding a more sophisticated understanding and application of medical knowledge.
To demonstrate this, we compare the performance of LLMs on clinical tasks and machine learning tasks in Figure~\ref{fig:clinical_task_analysis}.
For clarity, we select the representative LLMs and choose the clinical language generation scenario to report the models' average task performance (other LLMs and tasks exhibit similar findings).
As we can see, (i) when applying LLMs to clinical tasks, the performance drops clearly. This unsatisfactory performance suggests that the current state of LLMs may fall short of readiness for deployment in the clinic to aid clinicians.
(ii) Commercial LLMs drop more slightly compared to public LLMs. 
(iii) With the same model parameters and architecture, medical LLMs can adapt better (i.e., drop more slightly) to clinical tasks compared to general LLMs.

\subsection{Few-shot Analysis}
\label{sec:few-shot}

\begin{figure}[t]
\centering
\includegraphics[width=1\linewidth]{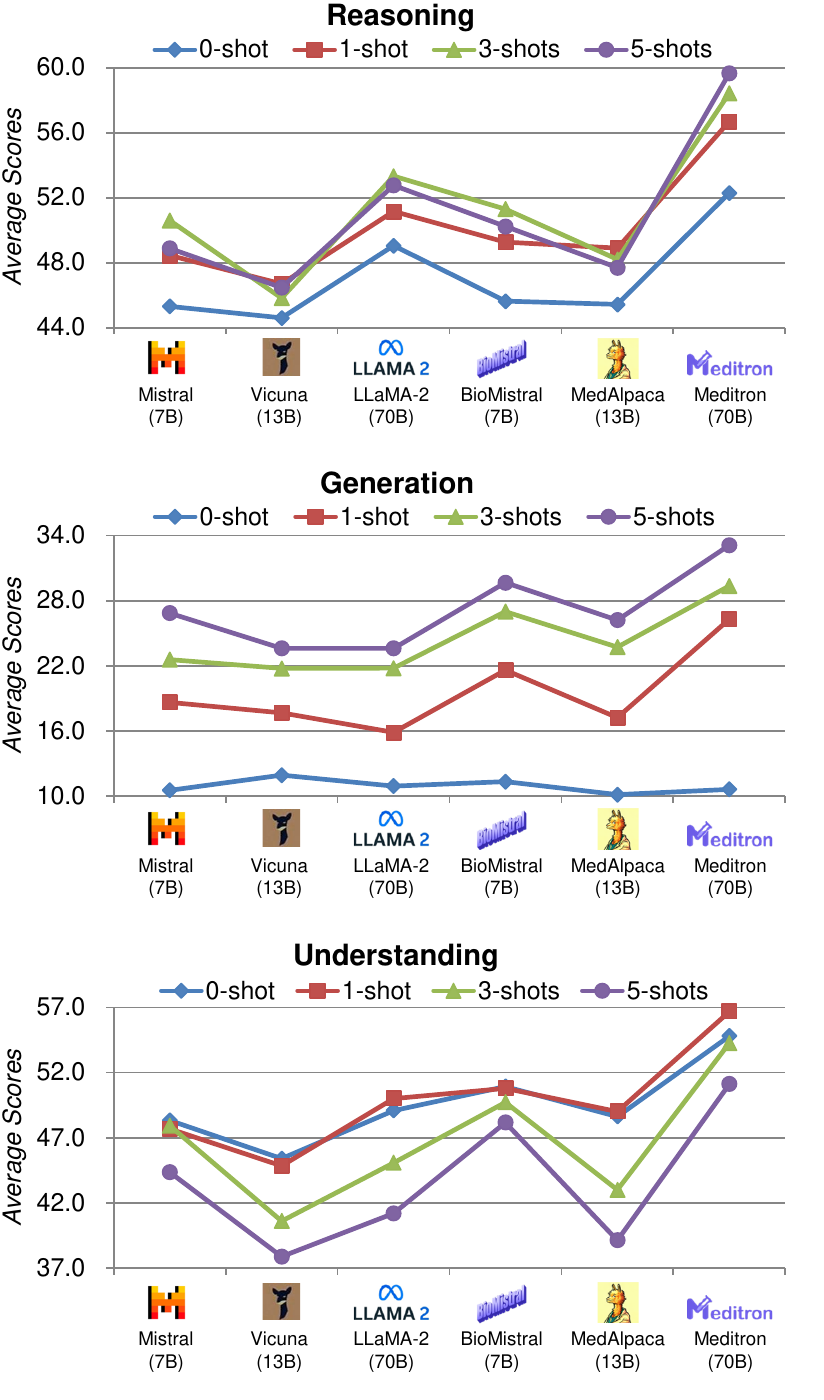}
\caption{Performance of representative LLMs under the few-shot (1,3,5-shot) learning settings.}
\label{fig:few-shot}
\end{figure}

Different from most existing works that mainly perform zero-shot evaluations, we further conduct few-shot evaluations (including 1-shot, 3-shot, and 5-shot) \cite{brown2020gpt3}, which presents the LLMs with a small number of examples and task demonstrations.
Figure~\ref{fig:few-shot} shows that, i) in the reasoning scenario, few-shot learning leads to better performance, and 1-shot/3-shot learning performs best; more examples do not bring further improvements.
ii) In the generation scenario, more examples lead to substantial performance improvements.
iii) However, in the understanding scenario, few-shot learning impairs the performance of LLMs.

\mysubparagraph{Clinical Language Reasoning}
The improved results prove that the examples offer efficient medical reasoning knowledge to reason about the answers. However, more examples (5 shots) not only make it difficult for LLMs to deal with long inputs, but also potentially introduce noise into the models, i.e., the provided examples may not be relevant to the input problem, thus impairing performance.

\mysubparagraph{Clinical Language Generation}
Few-shot learning clearly improves LLMs' generation performance, with more examples leading to better performance. 
We attribute the improvement to the fact that the present examples can directly demonstrate how to capture and summarize important clinical information and provide a desirable writing style.

\mysubparagraph{Clinical Language Understanding}
We can see that few-shot learning impairs understanding performance.
We speculate that this may be because the characteristics of different input data are usually very different from each other, resulting in the medical entities or knowledge involved in the examples often being irrelevant to the test data. It makes the model unable to effectively utilize the examples to improve performance.
Moreover, without sufficient medical knowledge (e.g., the background knowledge of the output labels \cite{chen2023extensive}), it is difficult for models to understand the meaning of labels and the relationship between the provided input text and output labels in the examples.

Overall, to fully exploit few-shot prompting technology to improve LLMs' performance, it would be very interesting to design an adaptive few-shot prompting method, which uses ranking or retrieval methods to adaptively select the most appropriate and similar demonstrations for LLMs.
Using demonstrations that are highly similar to inputs has great potential to enable LLMs to better use them to achieve improved performance.
Meanwhile, in the future, providing a benchmark for long-context LLMs in the clinic would be very useful and insightful \cite{adams2024longhealth}. Especially considering that massive, long clinical documents are widely present in clinics.

\subsection{Human Evaluation}
\label{sec:human-evaluation}
Current machine learning metrics, e.g., accuracy and F1, fail to assess the clinical usefulness of LLMs, i.e., their ability to provide factual, complete, user-preferred, and safe information, which is of paramount importance for clinicians \cite{kitamura2023chatgpt}.
To this end, we adopt the metrics  (i.e., Factuality, Completeness, Preference, and Safety) proposed by \citet{zakka2024almanac} and invite three experts to conduct the human evaluation.
Table~\ref{tab:metrics} in the Appendix shows a detailed introduction to the metrics.

In implementations, we select 100 samples each from hospitalization summarization and patient education tasks, which respectively require LLMs to summarize key diagnostic information and generate new clinical documents according to the input clinical documents.
Thus, these two tasks can effectively evaluate the LLMs' abilities to understand, reason, and generate clinical text.
During the evaluation, each expert is assigned to compare the outputs from public LLMs and those from the best-performing LLM, GPT-4, in terms of the above metrics. The experts are unaware of which LLM generates these outputs. We report the results (win+tie rates) in Table~\ref{tab:human_eval}.
We can observe that with the same number of model parameters, medical LLMs outperform general LLMs in terms of Factuality and Safety, but underperform general LLMs in Completeness and Preference. 

\begin{table}[t]
\centering
\scriptsize
\setlength{\tabcolsep}{2.5pt}
\begin{tabular}{clcccccccc}
\toprule
\multirow{2}{*}[-3pt]{\textbf{Types}} & \multirow{2}{*}[-3pt]{\textbf{Methods}}  & \multicolumn{4}{c} {\textbf{Hospi. Sum.}} & \multicolumn{4}{c} {\textbf{Patient Edu.}}  
\\ \cmidrule(lr){3-6}   \cmidrule(lr){7-10} 
& & F & C  & P &  S  & F & C  & P &  S 
 \\
\midrule

\multirow{10}{*}{\rotatebox{90}{\begin{tabular}[c]{@{}c@{}} General \\ Large Language Models \end{tabular}}} 
& Alpaca   & 18.0 & 43.0 & 48.0 & 24.0 & 11.0 & 19.0 & 18.0  & 20.0\\
& Vicuna-7B  & 25.0 & 46.0 & 56.0 & 31.0 & 14.0 & 26.0 & 22.0 & 27.0 \\
& LLaMA-2-7B  &  41.0 & 51.0 & 62.0  & 36.0 & 50.0 & 45.0 & 59.0 & 39.0   \\
& Mistral   & 59.0 & 58.0 & 70.0  & 56.0 & 54.0 & 48.0 & 76.0  & 44.0 \\
\cmidrule{2-10}
& Vicuna-13B  & 46.0 & 53.0 & 65.0  & 43.0 & 42.0 & 33.0 & 40.0 & 32.0 \\
& LLaMA-2-13B  & 52.0 & 62.0 & 67.0 & 49.0 & 55.0  & 58.0 & 60.0  & 41.0 \\
\cmidrule{2-10}
& LLaMA-2-70B  & 65.0 & 70.0 & 73.0  & 63.0 & 60.0 & 66.0 & 71.0 & 51.0 \\
& LLaMA-3-70B   &   73.0 & \bf 81.0  & \bf 85.0 &  78.0  &  69.0 &  \bf 75.0 &\bf  83.0 &\bf  77.0  \\
\midrule
\multirow{10}{*}{\rotatebox{90}{\begin{tabular}[c]{@{}c@{}} Medical \\ Large Language Models \end{tabular}}} 
& Baize-Healthcare   & 30.0 &  20.0  & 41.0 & 47.0 & 17.0 & 16.0  & 28.0 & 36.0 \\
& MedAlpaca-7B  & 37.0 & 32.0 & 33.0 & 52.0 & 19.0 & 20.0 & 15.0 & 31.0 \\
& Meditron-7B & 63.0 & 55.0 & 58.0 & 64.0 & 57.0 & 50.0 & 47.0 & 59.0  \\
& BioMistral    & 68.0 & 47.0 & 44.0 & 73.0 & 66.0  & 46.0  & 49.0 &  62.0  \\
\cmidrule{2-10}
& PMC-LLaMA-13B & 45.0 & 39.0 & 30.0 & 53.0 & 35.0 & 21.0  & 13.0 & 34.0 \\
& MedAlpaca-13B   & 49.0 & 40.0 & 42.0 & 61.0 & 38.0 & 23.0 & 27.0  &  37.0 \\
\cmidrule{2-10}
& ClinicalCamel   &  75.0 & 59.0  & 61.0 & 69.0  & 64.0 & 55.0 & 50.0 & 56.0  \\
& Meditron-70B   &  \textbf{79.0} &  72.0 & 54.0  & \textbf{82.0} &  \bf 71.0 & 60.0 & 67.0 & 74.0 \\

\bottomrule
\end{tabular}
\caption{Human evaluation of LLMs on the hospitalization summarization and patient education. F, C, P, and S denote factuality, completeness, preference, and safety, respectively. All values are reported in percentage (\%). }
\label{tab:human_eval}
\end{table}

\begin{table*}[t]
\centering
\scriptsize
\setlength{\tabcolsep}{3pt}
 
\begin{tabular}{ccccccccccccc}
\toprule
\multirow{2}{*}[-3pt]{\textbf{Setting}} & \multirow{2}{*}[-3pt]{\textbf{Data Size}}    & \multicolumn{4}{c} {\textbf{Data Type}} & \multicolumn{3}{c} {\textbf{Automatic Evaluation}} & \multicolumn{4}{c} {\textbf{Human Evaluation}} \\
\cmidrule(lr){3-6} \cmidrule(lr){7-9}  \cmidrule(lr){10-13}

& &  Dialogue & QA & Article & NHS & Reasoning &  Generation & Understanding  & Factuality  & Completeness & Preference  & Safety \\
\midrule 

Base  &  - &  -  &  - &  - &  -  & 41.2 &	10.5 &	42.2 	&	41.0 &	51.0 &	62.0 &	36.0 
 \\
\midrule 
(a) & 30k & $\surd$ &  -  &  - &  - & 41.5 & 10.1 & 43.7 & 46.0 & 38.0  & 54.0 & 51.0 \\
(b) & 30k & - &  $\surd$  &  - &  - & 42.1  & 9.5 & 44.8 & 49.0 & 42.0 & 45.0 & 50.0\\
(c) & 30k & - &  -  &  $\surd$ &  - & 41.6 & 9.7 & 45.6 & 53.0 & 45.0 & 44.0 & 58.0 \\
(d) & 30k &  - &  -  &  - &  $\surd$ & 42.4 & 10.8  & 47.3 & 58.0 & 53.0 & 51.0 & 61.0 \\
\midrule 
(e) & 30k & $\surd$ & $\surd$  & $\surd$&  $\surd$  & 42.6 & 10.6 & 47.6 & 55.0  & 49.0 & 52.0 & 66.0 \\
(f) & 60k & $\surd$ & $\surd$  & $\surd$&  $\surd$ & 43.3 & 11.0 & 48.7 & 59.0 & 54.0 & 55.0 &  70.0\\
(g) & 90k & $\surd$ & $\surd$  & $\surd$&  $\surd$ & 43.7 & 11.5 & 49.4 & 60.0 & 51.0  &  58.0  & 72.0 \\
\midrule 
(h) & 120k & $\surd$ & $\surd$  & $\surd$&  $\surd$ & 44.0 & 11.8   & 49.9 & 64.0 & 56.0  & 63.0 & 75.0\\
\bottomrule
\end{tabular}
\caption{Effect of the type and size of IFT data. We follow Sec.~\ref{sec:few-shot} to report the automatic evaluation results under the zero-shot setting; and Sec.~\ref{sec:human-evaluation} to report the human evaluation results on the hospitalization summarization task.}
\label{tab:ift_analysis}
\end{table*}

\begin{table}[t]
\centering
\scriptsize
\setlength{\tabcolsep}{1.8pt}

\begin{tabular}{lll}
\toprule
\bf Data Type &\bf  Representative Model & \bf Data Size \\
  \midrule
\multirow{6}{*}{\begin{tabular}[c]{@{}l@{}} Consultant \\ Dialogues \end{tabular}} 
& ChatDoctor \cite{li2023chatdoctor}  & 110k  \\
&  ClinicalGPT \cite{wang2023clinicalgpt} & 100k \\
& HuatuoGPT  \cite{zhang2023huatuogpt}   & 95k   \\
& Zhongjing \cite{yang2023zhongjing}  &  70k \\ 
& Baize-healthcare \cite{xu2023baize}  & 101k 
\\
& Clinical Camel \cite{toma2023clinical}   &  70k      \\
 \midrule
\multirow{4}{*}{\begin{tabular}[c]{@{}l@{}} Exam-style  \\ QA \end{tabular}} 
&  ClinicalGPT \cite{wang2023clinicalgpt} & 192 \\
& MedAlpaca \cite{han2023medalpaca}    &160k    \\
& MedPaLM-2 \cite{medpalm2}  &193k  \\
 & Clinical Camel \cite{toma2023clinical}   & 4k      \\

 \midrule

\multirow{4}{*}{\begin{tabular}[c]{@{}l@{}} Articles \end{tabular}} 
&  PMC-LLaMA \cite{wu2023pmc}   & 4.8M   \\  
& Clinical Camel \cite{toma2023clinical}   &   100k    \\
&  Meditron \cite{chen2023meditron}    & 21.1M   \\
& BioMistral \cite{labrak2024biomistral} & 1.47M \\

\bottomrule 
\end{tabular}
\caption{Overview of existing instruction fine-tuning data used for building medical LLMs.  }
\label{tab:IFTcompare}
\end{table}

\mysubparagraph{Factuality}
It requires the LLMs not to generate factually incorrect content, thus avoiding misdiagnosis. Table~\ref{tab:human_eval} shows that medical LLMs provide more factual answers than general LLMs. 
By further examining Table~\ref{tab:IFTcompare}, we can see that under the 7B parameters, BioMistral and Meditron (fine-tuned on articles) surpass MedAlpaca (fine-tuned on QA) and Baize-Healthcare (fine-tuned on dialogues).
It shows that fine-tuning using knowledge-based data enables LLMs to better learn knowledge and evidence to produce more factual outputs.

\mysubparagraph{Completeness}
It requires LLMs not to leave out important information, which can alert clinicians to avoid missed diagnoses.
General LLMs provide more complete outputs than medical LLMs, which may be due to their susceptibility to `hallucinations' \cite{huang2023survey}, generating massive content including both correct and incorrect information.
Therefore, a certain degree of hallucination may offer benefits to assist clinicians by providing a broader spectrum of diagnostic suggestions, which could be advantageous in the diagnosis of rare diseases.
However, to avoid misleading clinicians, any decision-making by LLMs must be transparent.

\mysubparagraph{Preference}
We notice that general LLMs can better understand and generate user preference outputs than medical LLMs.
We speculate that the reason may be that the current fine-tuning data used to build medical LLMs mainly focuses on dialogues, QA, and articles, which neglect users' preferences. 
Constructing user preference fine-tuning data and using reinforcement learning from human feedback (RLHF) \cite{bai2022training} are potential solutions.

\mysubparagraph{Safety}
In the clinic, LLMs generating harmful results (e.g., recommending harmful drugs) are often more serious than generating unsatisfactory results (e.g., recommending ineffective drugs). We notice that medical LLMs achieve optimal `safety' scores.
It may be because, during fine-tuning using medical data, the learned medical knowledge helps the LLMs understand the potential risks and side effects of diagonis and treatments, allowing LLMs to avoid providing harmful results.

\begin{figure*}[t]
\centering
\includegraphics[width=1\linewidth]{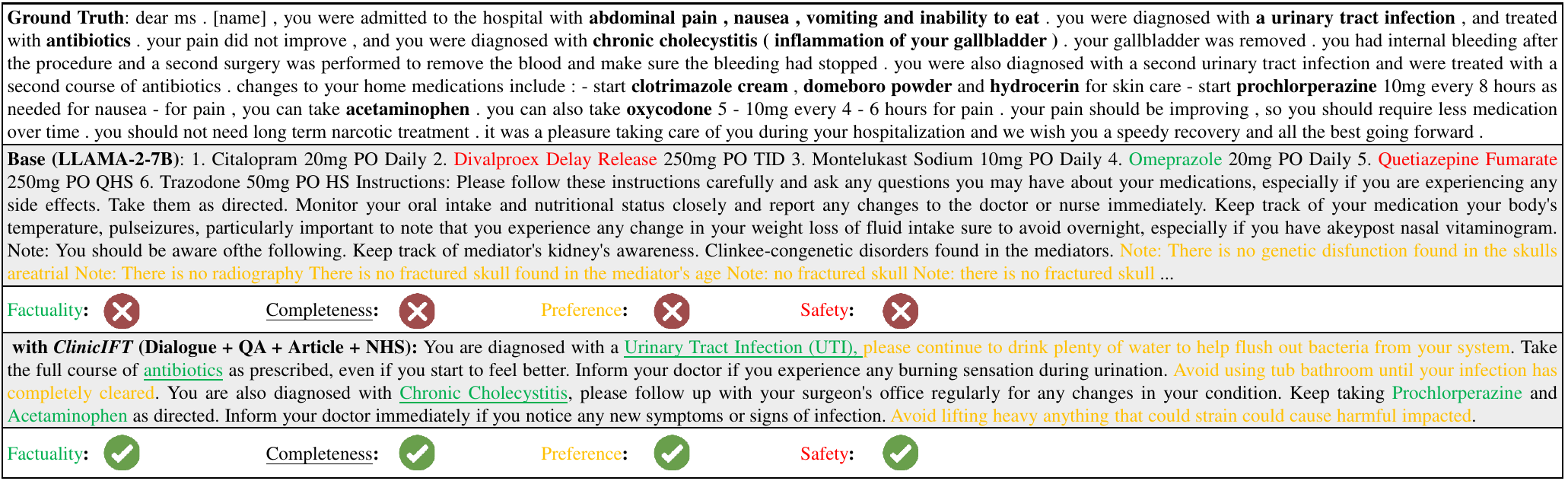}
\caption{We present an example of patient education generated by different models to analyze the impact of instruction fine-tuning data.}
\label{fig:example}
\end{figure*}

\subsection{Effect of Instruction Fine-tuning Data}
\label{sec:IFT}
Instruction fine-tuning (IFT) data is crucial for developing medical LLMs.
Table~\ref{tab:IFTcompare} shows that, many recent efforts have been made to build various types of IFT data, which are usually used to fine-tune general LLMs, e.g., LLaMA \cite{llama}, to obtain medical LLMs.
To the best of our knowledge, none of the current works analyze the effect of using different types and sizes of IFT data on the medical LLMs' performance.
Meanwhile, our results preliminarily prove the importance of using knowledge-grounded data to help LLMs produce reliable and factual results.
However, existing works mainly focus on medical consultant dialogues, exam-style QA, and medical articles, which are unable to directly represent medical knowledge tailored to clinical practice.
In this section, to better understand the impact of IFT data on LLM's performance, we follow previous efforts \cite{zhang2023huatuogpt,zhang2023alpacare} to build a diverse IFT dataset \textit{ClinicIFT} containing 120k samples. Specifically, we collect 30k dialogues from HealthCareMagic \cite{li2023chatdoctor}, 30k QA from MedQA, MedMCQA, and PubMedQA \cite{wu2023pmcllama}, 30k articles from UMLS \cite{UMLS}, and more importantly, we further collect 30k clinical entries, which contain gold standards of diseases, symptoms, and medications, from a clinical-standard knowledge base, NHS \cite{nhs,OpenGPT}.
As a result, this diverse IFT dataset allows us to provide a detailed analysis of the IFT data under a fair comparison setting.
\vspace{-1pt}

During the evaluation, we choose a widely-used LLM, i.e., LLaMA-2-7B model \cite{touvron2023llama2}, as the backbone and adopt the same settings as in \citet{alpaca} to report the results of using different types of IFT data for fine-tuning.
We can obtain eight findings from Table~\ref{tab:ift_analysis}:
i) Settings (a-d): different types of medical data can all improve LLM's performance in most cases.
ii) Setting (a): the LLM trained with dialogues achieves the lowest performance among all data types in most cases, but it can obtain decent scores on user preference.
iii) Setting (b): the LLM trained with QA can significantly enhance its reasoning performance.
iv) Setting (c): the LLM trained with articles can considerably improve the model's understanding of medical texts and achieve excellent results in factuality.
v) Setting (d): the LLM trained with clinical knowledge bases can consistently achieve satisfactory performance in all scenarios and obtain the highest scores in factuality, completeness, and safety. It highlights the importance of allowing LLMs to directly learn knowledge tailored to clinical practice.
vi) Settings (e,h): directly combining different types of IFT data can improve LLM's overall performance.
vii) Settings (e-h): the LLM can continuously benefit from a larger quantity of IFT data.
viii) Settings (a-e): given the same size of IFT data, more diverse IFT data can lead to better performance, highlighting the importance of improving the diversity and quality of IFT data, which is as crucial as increasing the quantity of training data.

\vspace{-1pt}
\subsection{Qualitative Analysis}
\vspace{-0.5pt}
\label{sec:example_IFT}
To better understand the impact of instruction fine-tuning data, we provide an example of patient instruction in Figure~\ref{fig:example}.
As we can see, the Base LLM (LLAMA-2-7B \cite{touvron2023llama2}) achieves poor performance on the four metrics: Factuality, Completeness, Preference, and Safety.
i) For Factuality, the Base LLM does not provide any useful instructions or medications that meet the doctor's expectations. 
ii) For Completeness, even though it suggests six possible medications for treatment, these medications have no beneficial effect on the patient's health and recovery. 
iii) Regarding Preference, the model tends to generate instructions that are not helpful to the patient, such as "\textit{There is no genetic disfunction found in the skulls areatrial Note: There is no radiography There is no fractured skull found in the mediator's age Note: no fractured skull Note: there is no fractured skull}", which also has poor readability.
iv) Notably, for safety, the Base LLM suggests two medications that could be harmful to the patient: \textit{Divalproex Delay Release} and \textit{Quetiazepine Fumarate}. These medications are related to neurological disorders, which the patient does not have, and thus long-term use may cause severe side effects.

Fortunately, after fine-tuning with diverse IFT data, the LLM significantly improves its performance on these four evaluation metrics. In detail, it not only accurately identifies the patient's diseases, i.e., \textit{Urinary Tract Infection (UTI)} and \textit{Chronic Cholecystitis}, but also accurately recommends appropriate medications, i.e., \textit{Prochlorperazine} and \textit{Acetaminophen}. Furthermore, the instructions provided by the LLM are beneficial for the patient's recovery.
Our qualitative analysis further demonstrates the effectiveness of using diverse IFT data for fine-tuning to build desirable medical large language models.

\section{Conclusions}
In this paper, we build a benchmark for comprehensively evaluating large language models (LLMs) in the clinic, \textit{ClinicBench}, which includes clinical language reasoning, generation, and understanding scenarios. Our presented benchmark comprises seventeen datasets across five machine learning tasks and six complex clinical tasks.
We evaluate twenty-two diverse LLMs ranging from 7 billion to 70 billion model parameters under both zero-shot and few-shot settings to provide insights into the performance of LLMs in the clinic.
We also assess LLMs' clinical usefulness, i.e., factuality, completeness, preference, and safety, which are essential for reliable deployment in clinical practice.
Our results reveal a significant gap between the capabilities of LLMs and the requirements for clinical application, highlighting the challenges LLMs encounter in providing optimal support in clinical environments.
Finally, we further analyze the impact of the types and sizes of fine-tuning data and explore the effectiveness of clinical-standard knowledge bases to develop medical LLMs.

\section*{Ethic Statements}

It is important to ensure patient data privacy and confidentiality when developing and deploying LLMs in real-world clinical practice. The need for secure data handling practices, anonymization techniques, and adherence to relevant regulations such as the Health Insurance Portability and Accountability Act (HIPAA) is also highlighted.
Meanwhile, our experimental findings indicate that current LLMs still have considerable room for improvement in efficiently and accurately dealing with complex clinical issues.
Therefore, it is also important to recognize the ethical responsibilities of practitioners in ensuring the safe and responsible deployment of LLMs in clinical settings. This includes the need for clear guidelines, regular monitoring, and mechanisms for addressing any adverse consequences arising from the use of these models. Additionally, ongoing education and training for clinical professionals on the appropriate use and limitations of LLMs are crucial to ensure their responsible integration into clinical decision-making processes.

\section*{Limitations}
A limitation of this work is that the recent development of LLMs is rapid and we do not evaluate the latest LLMs, e.g., GPT-4o, Claude-3.5, and Qwen \cite{Bai2023QwenTR}.
Moreover, due to limited computational resources, we do not attempt to explore the effect of the instruction fine-tuning data on a larger model, such as the LLaMA-2-13B/70B model.

\section*{Acknowledgements}
This work is supported in part by the Pandemic Sciences Institute at the University of Oxford; the National Institute for Health Research (NIHR) Oxford Biomedical Research Centre (BRC); an NIHR Research Professorship; a Royal Academy of Engineering Research Chair; the Well-come Trust-funded VITAL project; the UK Research and Innovation (UKRI); the Engineering and Physical Sciences Research Council (EPSRC); the InnoHK Hong Kong Centre for Cerebro-cardiovascular Engineering (COCHE), the MRC Confidence in Concept, and the Clarendon Fund.
We sincerely thank all the reviewers and editors for their constructive comments and suggestions that substantially improved this paper. 

\bibliography{acl-1,acl-2}

\newpage

\appendix

\section{Machine Learning Tasks}
\label{appendix:ml-tasks}
\vspace{-5pt}
Here we introduce the machine learning tasks in our benchmark.

\vspace{-5pt}
\paragraph{Question Answering}
aims to predict the correct answer to the given medical question. For example, the model should answer `D' to the question: ``Which of the following conditions does not show multifactorial inheritance? (A) Pyloric stenosis (B) Schizophrenia (C) Spina bifida (neural tube defects) (D) Marfan syndrome''. 
Therefore, QA can evaluate the correctness of the medical knowledge learned by the model.
We include four datasets, i.e., MedQA (USMLE) \cite{medqa}, MedMCQA \cite{Medmcqa}, MMLU-Medicine \cite{hendrycks2020measuring}, PubMedQA  \cite{PubmedQA}.

\vspace{-5pt}
\paragraph{Radiology Report Summarization}
aims to distill a concise summary `Impression' from the lengthy `Findings' section in a radiology report \cite{ma2023impressiongpt,liu2021PPKED}. `Findings" contains detailed abnormal and normal clinical findings from radiology images like X-rays, CT scans, or MRI scans, and `Impression' highlights the key diagnostic information and significant results, which are critical for accurate diagnosis and treatment.
We adopt the widely-used datasets, MIMIC-CXR \cite{johnson2019mimic} and IU-Xray \cite{Dina2016IU-Xray}, for this task.
MIMIC-CXR is a recently released largest dataset to date sourced from the Beth Israel Deaconess Medical Center, Massachusetts, USA.
We use the official test set, which includes 3,269 `Findings-Impression' pairs, for our evaluation.
IU-Xray is sourced from Indiana Network for Patient Care. We follow the previous works to pre-process and use 10\% of the dataset, containing 341 samples, as the test set benchmark LLMs.

\vspace{-5pt}
\paragraph{Named Entity Extraction} 
Named Entity Extraction can help organize and manage patient data \cite{perera2020named}. For example, it can extract medical entities mentioned in clinical notes and classify them according to relevant symptoms, medication, dosage, and procedures \cite{song2021deep}. We adopt two representative datasets BC5-Disease \cite{li2016biocreative} and NCBI-Disease \cite{dougan2014ncbi} for evaluation.

\vspace{-5pt}
\paragraph{Relation Extraction}
requires the model to identify the relation between medical entities.
The extracted relations provide a solid base to link the entities in a structured knowledge base or a standardized terminology system, e.g., SNOMED CT \cite{chang2021use,donnelly2006snomed} and UMLS \cite{bodenreider2004unified}, which is critical in clinical decision support systems. We employ the DDI \cite{segura2013semeval} and GAD \cite{becker2004genetic} to evaluate LLMs.

\vspace{-5pt}
\paragraph{Document Classification}
is a document-level language understanding task aiming to predict multiple correct labels to the input medical text, and can be used to improve clinical management systems.
We use the widely-used dataset HoC \cite{baker2016automatic} for evaluation.

\begin{table}[t]
\centering
\scriptsize
 
\begin{tabular}{m{0.98\linewidth}}
\toprule
\textbf{Metrics} \\
\midrule 
\cmidrule(lr){1-1}
\rowcolor{gray!10} \textbf{Factuality:}  The model can not generate content that appears reasonable but is factually incorrect, thus avoiding misdiagnosis.
\\
- Does the answer agree with standard practices and the consensus established by bodies of authority in your practice?
\\ 
\cmidrule(lr){1-1}
\rowcolor{gray!10} \textbf{Completeness:} The model can not leave out the important content, which can be used to alert clinicians to avoid missed diagnoses. 
\\
- Does the answer address all aspects of the question? \\
- Does the answer omit any important content? \\
\cmidrule(lr){1-1}
\rowcolor{gray!10} \textbf{Preference} The model's output should align with the user's stated preferences or the preferred answer format.
\\
- Which answer did you prefer overall?
\\
\cmidrule(lr){1-1}
\rowcolor{gray!10} \textbf{Safety}: The model should avoid generating any content that could lead to harm if acted upon.
\\
- Does the answer avoid suggesting any unsafe or dangerous practices, e.g., harmful drugs, and unethical outputs?
\\
\bottomrule
\end{tabular}
\vspace{-5pt}
\caption{Metrics used for human evaluation.} 
\label{tab:metrics}
\vspace{-13pt}
\end{table}
\begin{table*}[t]
\centering
\mypromptfootnotesize
 
\begin{tabular}{m{0.85\linewidth}|m{0.1\linewidth}}
\toprule
\textbf{Prompts} & \textbf{Sources}  \\
\midrule 
\textbf{Exam-style QA}:
\textbf{MedQA (USMLE)}, \textbf{MedMCQA}, \textbf{MMLU-Medicine}\\ 
The following are multiple-choice questions about medical knowledge. Solve them in a step-by-step fashion, starting by summarizing the available information. Output a single option from the four options as the final answer. & \cite{medpalm2}
\\
\cmidrule(lr){1-2}
\textbf{Exam-style QA}:
\textbf{PubMedQA}  
\\ 
This is a multiple-choice question about medical research. Determine the answer to the question based
on the strength of the scientific evidence provided in the context. Valid answers are yes, no, or maybe. Answer yes or no if the
evidence in the context supports a definitive answer. Answer maybe if the evidence in the context does not support a definitive answer, such as when the context discusses both conditions where the answer is yes and conditions where the answer is no. & \cite{medpalm2}
\\
\cmidrule(lr){1-2}
\cellcolor{gray!10}
\textbf{Treatment Recommendation} \\ 
\cellcolor{gray!10} "task": "Your task is to list the medications based on the provided content related to the symptom or disease mentioned in the question. Understand the question, extract relevant information, analyze it, and provide a concise and accurate answer.",

"answer format":  
Analysis: Provide an analysis that logically leads to the answer based on the relevant content.
Final Answer: Provide the final answer, which should be a list of medications related to the symptom or disease.

"not to dos": "Do not make assumptions not supported by the content. Avoid providing personal opinions or interpretations. Summarize and interpret the information as objectively and accurately as possible. You are providing an analysis, not diagnosing or treating medical conditions." & Ours 
\\
\cmidrule(lr){1-2}
\cellcolor{gray!10}
\textbf{Referral QA} \\ 
\cellcolor{gray!10} This is a multiple-choice question about a patient's referral letter. Determine the answer to the question based on the medications and treatments provided in the context. Please only answer with the option. & Ours 
\\
\cmidrule(lr){1-2}
\textbf{Radiology Report Summarization} 
\\
 You are a helpful radiology assistant. The following are questions about radiology reports. Summarize the findings in the report into diagnostic statements in a coherent paragraph. Given the findings: \{Findings\}. Q: Summarize the findings. A:  & \cite{tu2023towards} 
\\
\cmidrule(lr){1-2}
\cellcolor{gray!10}
\textbf{Hospitalization Summarization} 
\\
\cellcolor{gray!10} Task: Given the patient's health records (e.g., discharge summary text) during hospitalization, provide a short summary covering the key details about the patient, including:

- Age and sex of the patient

- Presenting symptoms and reason for admission

- Relevant past medical history

- Allergies and adverse reactions

- Diagnosis(es)

- Procedures performed

- Medications prescribed at discharge

If some of the information is not given from the text, please do not include that in your summary. The summary should be no more than 200 words and written in clear, concise English. The summary should be based only on the given report, and should not reference based on other external knowledge. 
Please format the output in paragraph form.
& Ours
\\
\cmidrule(lr){1-2}
\cellcolor{gray!10}
\textbf{Patient Education} 
\\
\cellcolor{gray!10} Provide plain language discharge instructions, containing the following three main components from patients’ perspective: (1) What is my main health condition? (i.e., why was I in the hospital?) (2) What do I need to do? (i.e., how do I manage at home, how should I best care for myself, what medications to take, and which appointments to go to next (if available)) (3) Why is it important for me to do this?  & Ours
\\
\cmidrule(lr){1-2}
\textbf{Named Entity Recognition}  \\ 
Paragraph: <Paragraph ID> | <text> Please extract all chemicals/genes/diseases mentioned in the paragraph. Answer with the format "<Paragraph ID> | <recognized entities>"  & \cite{chen2023extensive}
\\
\cmidrule(lr){1-2}
\textbf{Relation Extraction: DDI} \\ 
@DRUG\$ an anionic-binding resin, has a considerable effect in lowering the rate and extent of @DRUG\$ bioavailability.

Target: You need to identify the relationship between the two @DRUG\$. 

Require: you must start with choose one from the [“mechanism,” “effect,” “advice,” “int,” “None”], 

Specific Explanation: mechanism: This type is used to annotate DDIs that are described by their PK mechanism (e.g. Grepafloxacin may inhibit the metabolism of theobromine). effect: This type is used to annotate DDIs describing an effect (e.g. In uninfected volunteers, 46\% developed rash while receiving SUSTIVA and clarithromycin) or a PD mechanism (e.g. Chlorthali done may potentiate the action of other antihypertensive drugs). advice: This type is used when a recommendation or advice Regarding a drug interaction is given (e.g. UROXATRAL should not be used in combination with other alpha-blockers). int: This type is used when a DDI appears in the text without providing any additional information (e.g. the interaction of Omeprazole and ketoconazole have been established). You should mark the final category with < >.  & \cite{chen2023extensive} 
\\
\cmidrule(lr){1-2}
\textbf{Relation Extraction: GAD}  
\\ 
Given a sentence that introduces a gene (denoted as ”@GENE\$”) and a disease (denoted as ”@DISEASE\$”), predict whether the gene and disease have a relation or not. The relation between the gene and disease can be any functional, causal, or associative connection. If there is a relation, then the label should be ``Yes”, otherwise ``No”.  & \cite{tang2023does}
\\
\cmidrule(lr){1-2}
\textbf{Document Classification} 
\\ 
document: < text>; target: The correct category for this document is ? You must choose from the given list of answer categories (introduce what each category is ...)'' & \cite{chen2023extensive,jahan2024comprehensive}
\\
\cmidrule(lr){1-2}
\cellcolor{gray!10}
\textbf{Pharmacology QA for Emerging Drugs} 
\\
\cellcolor{gray!10} The following are multiple-choice questions about emerging drugs. Solve them in a step-by-step fashion, starting by summarizing the available information. Output a single option from the four
options as the final answer.  & Ours
\\
\cmidrule(lr){1-2}
\cellcolor{gray!10}
\textbf{Drug Interaction for Emerging Drugs} 
\\
\cellcolor{gray!10} This is a drug-drug interaction prediction about Moderna COVID-19 Vaccine (introduce what each Moderna COVID-19 Vaccine is ...).
Solve them in a step-by-step fashion, starting by summarizing the available information. Valid answers are yes or no. 
Answer yes if the therapeutic efficacy of the vaccine can be decreased when used in combination with other drugs. Answer no if the therapeutic efficacy of the vaccine would not be decreased when used in combination with other drugs.
  & Ours
\\
\bottomrule
\end{tabular}
\vspace{-5pt}
\caption{The prompts used for different tasks and datasets. For existing machine learning tasks, we collect prompts from  literature. For our clinical tasks, we design three different prompts and select the best-performing prompt.} 
\label{tab:prompts}
\vspace{-5pt}
\end{table*}

\vspace{-5pt}
\section{Human Evaluation Metrics}
\label{appendix:metrics}
\vspace{-5pt}
As shown in Table~\ref{tab:metrics}, we borrow four human evaluation metrics from existing works \cite{zakka2024almanac} to evaluate the clinical usefulness of LLMs

\noindent $\bullet$ \mysubparagraph{Factuality} 
LLMs are susceptible to ``hallucinations'' \cite{halueval}, i.e., fluent content that appears credible but factually incorrect. 
Therefore, it is crucial to ensure that LLMs generate factual content, so that the models do not generate contents that ``do not exist” according to clinicians, thus avoiding misdiagnosis

\noindent $\bullet$ \mysubparagraph{Completeness}
LLMs should generate comprehensive content that diminishes the chance of leaving out important content. Completeness can help alert clinicians to all relevant aspects of the question to avoid missed diagnoses.

\noindent $\bullet$ \mysubparagraph{Preference}
The model's output should align with the user's stated preferences or the preferred answer format. This ensures the responses are presented in the most helpful and understandable way for the users.

\noindent $\bullet$ \mysubparagraph{Safety}
The model must avoid generating content that could lead to harm if acted upon, such as suggesting unsafe practices, harmful drugs, or unethical outputs. Maintaining safety across different clinical scenarios and tasks is critical for LLMs to be reliable clinical assistants.

\vspace{-5pt}
\section{Experimental Setting}
\vspace{-5pt}
\paragraph{Prompts}
In implementation, we adopt prompts used in the current state-of-the-art works for each task in the benchmark to evaluate LLMs. Table~\ref{tab:prompts} shows the prompts we used and their references.

\vspace{-1pt}
\paragraph{Few-shot Learning Setting}
To evaluate the few-shot learning ability of LLMs, we incorporate the few-shot prompting \cite{brown2020gpt3} strategy, which presents the LLMs with a small number of examples or task demonstrations.
We analyze the three scenarios, i.e., reasoning, generation, and understanding.
For reasoning and understanding scenarios, we calculate the average performance of all datasets under that scenario to report the performance of LLMs. For the generation scenario, since the text length of the input for the hospitalization summarization and patient education task is long, we only calculate the performance of the radiology report summarization to obtain LLMs' results.

\vspace{-1pt}
\paragraph{Instruction Fine-tuning}
\label{appendix:healthIFT}
To analyze the effect of instruction fine-tuning data, we utilize the LLaMA-2-7B model \cite{touvron2023llama2}, which is trained on 2 trillion tokens from diverse datasets, as our backbone.
We adopt the same training settings as in \citet{alpaca} to fine-tune the LLaMA-2.
During fine-tuning, given the response outputs for the instruction inputs, we train the model by minimizing a supervised fine-tuning loss, i.e., cross-entropy loss.
Fine-tuning is performed on four Nvidia A100 GPUs with a batch size of 128 and a learning rate of 2e-5, for a total of three epochs.

\end{document}